\documentclass[conference,onecolumn]{IEEEtran}
\IEEEoverridecommandlockouts
\usepackage{cite}
\usepackage{amsmath,amssymb,amsfonts}
\usepackage{graphicx, float}
\usepackage{textcomp}
\usepackage{xcolor}
\usepackage{graphics}
\usepackage{enumitem}

\usepackage{csquotes}
\usepackage{subcaption}
\usepackage{multicol}
\usepackage{lipsum}
\usepackage[bookmarks=false]{hyperref}
\usepackage{amsmath,amssymb}
\usepackage{amsfonts}
\usepackage{float}
\usepackage{wrapfig}
\def\BibTeX{{\rm B\kern-.05em{\sc i\kern-.025em b}\kern-.08em
    T\kern-.1667em\lower.7ex\hbox{E}\kern-.125emX}}
    
\usepackage{tabularx}

\usepackage{algorithm}
\usepackage{algpseudocode}
\algblock{Input}{EndInput}
\algnotext{EndInput}
\algblock{Output}{EndOutput}
\algnotext{EndOutput}
\usepackage{authblk}

\usepackage{booktabs}
\usepackage{multirow}

\usepackage{tikz}
\usetikzlibrary{matrix}
\usepackage{xparse}
\usepackage{soul}

\ExplSyntaxOn
\NewDocumentCommand{\avercalc}{m}{%
  \clist_set:Nn \l_tmpa_clist {#1}%
  \fp_zero:N \l_tmpa_fp
  \clist_map_inline:Nn \l_tmpa_clist {
    \fp_add:Nn \l_tmpa_fp {##1}
  }
  \fp_eval:n { \l_tmpa_fp / \clist_count:N \l_tmpa_clist }
}
\ExplSyntaxOff

\usepackage{float} 
\newcommand*{\affaddr}[1]{#1} 
\newcommand*{\affmark}[1][*]{\textsuperscript{#1}}
\newcommand*{\email}[1]{\textit{#1}}

\begin{document}

\title{Mitigating The Effect of Class Imbalance in Data with Hierarchical and Dependable Structure}

\author{Bipin Chhetri\affmark[1],
 Deepika Giri\affmark[2], Avishek Kadel\affmark[3],  Rabin Kumar Karki\affmark[4], Akbar Siami Namin\affmark[1]\\
\affaddr{Department of Computer Science\affmark[1]}, 
\affaddr{Texas Tech University\affmark[1]} \\
\affaddr{Cumberland University\affmark[2]}, 
\affaddr{Yeshiva University\affmark[3]}, 
\affaddr{University of Cumberlands\affmark[4]}\\
\email{\{bipin.chhetri, akbar.namin\IEEEauthorrefmark{2}\}}@ttu.edu, 
 \email{dgiri25@students.cumberland.edu}, \\ \email{akadel@mail.yu.edu},
 \email{rkarki34351@ucumberlands.edu}
}

\maketitle

\begin{abstract}
Classifying cybersecurity vulnerabilities using the Common Weakness Enumeration (CWE) taxonomy is challenging due to extreme class imbalance and strong hierarchical dependencies among weakness categories. Although oversampling techniques such as Synthetic Minority Oversampling Technique (SMOTE) and Adaptive Synthetic Sampling (ADASYN) are widely adopted to mitigate class imbalance, their effectiveness for hierarchical CWE text classification remains largely unexplored. This paper proposes a Hierarchy-Aware RoBERTa framework that explicitly incorporates CWE structural information through learnable parent-class embeddings, preserving taxonomic consistency. Our experiments demonstrate that synthetic interpolation in high-dimensional embedding spaces violates the inherent parent-child constraints of the CWE hierarchy, offering only marginal benefits for classical ML models while consistently degrading deep learning architectures. Evaluated on a CWE Research Concept dataset, the proposed model achieves a weighted F1-score of $0.76$ without data augmentation, outperforming all baselines with notable gains on minority classes, including the \texttt{Class} category whose F1-score improves from $0.49$ to $0.60$ over the BERT baseline. Our results suggest that hierarchy-aware representation learning is a more principled alternative to oversampling for structured vulnerability classification.

\end{abstract}


\section{Introduction}

Mitigating weaknesses has become increasingly necessary as software systems grow in complexity and adversaries adopt sophisticated attack techniques. Software and hardware weaknesses cataloged in the CWE\cite{MITRE2025} enable systematic vulnerability assessment, mitigation planning, and security decision making\cite{chhetri2025application}. The CWE taxonomy organizes weaknesses into abstraction levels such as \texttt{Base}, \texttt{Class}, and \texttt{Pillar}, where upper-level nodes capture broad concepts, and lower-level nodes represent specific details. In practice, a small number of weakness types dominate the corpus, while many rare but high-impact categories remain significantly underrepresented.

Transformer-based encoders such as BERT \cite{devlin2019bert} and RoBERTa \cite{liu2019roberta} have become strong baselines for CWE text classification, capturing bidirectional contextual dependencies that outperform earlier Convolutional Neural Network (CNN) \cite{yamashita2018convolutional} and Recurrent Neural Network (RNN) architectures \cite{chung2014empirical}. However, class imbalance remains a fundamental challenge that degrades performance on rare CWE categories, and CWE labels carry explicit parent-child dependencies \cite{jiang2025hierarchy}. For instance, a \texttt{Variant} is always a child of a \texttt{Base}, meaning misclassification at higher abstraction levels cascades downward to finer-grained classes.

Research on hierarchical classification emphasizes that maintaining parent-child relationships is crucial for building reliable prediction models \cite{jiang2025hierarchy}. Traditional approaches, such as random oversampling and undersampling \cite{yang2024impact}, rebalance label distributions but risk overfitting on minority samples or losing valuable information from the majority class. More advanced methods, SMOTE \cite{chawla2002smote} and ADASYN \cite{he2008adasyn}, generate synthetic minority samples in feature space. SMOTE interpolates between minority instances to reduce classifier bias toward majority labels, while ADASYN concentrates synthesis on harder boundary regions where misclassifications are likely. The specific contributions of our work are as follows:

\begin{enumerate}
    \item We empirically evaluate ML (RF, SVM), deep learning (CNN, BiGRU), and transformer-based (BERT) models under SMOTE and ADASYN  resampling on an imbalanced hierarchical CWE dataset.
    \item We demonstrate that synthetic interpolation in high-dimensional embedding spaces violates CWE parent-child constraints, marginally helping classical ML models while consistently degrading deep learning models.
    \item We introduce a Hierarchy-Aware RoBERTa model that injects CWE structural priors through learnable parent-class embeddings for end-to-end joint modeling of semantic content and hierarchical structure.
    \item Our proposed model achieves a weighted F1-score of $0.76$ without augmentation, outperforming all baselines with \texttt{Class} F1 improving from $0.49$ to $0.60$ over BERT.
\end{enumerate}

This paper is structured into the following sections. Section~\ref{sec:related} presents the related work. Section~\ref{sec:technical} presents the technical background of the models. Sections \ref{sec:methodology} and \ref{sec:experiment} contain the methodology and experimental setup, respectively. 
Section \ref{sec:results} presents the results. Section~\ref{sec:discussion} presents the discussion and limitations. Section~\ref{sec:conclusion} concludes the paper with future work.

\section{Related Work}
\label{sec:related}

\subsection{CWE-Based Vulnerability Classification}
Prior work primarily treats CWE classification as a supervised multiclass problem in data-rich settings. Chhetri et al. \cite{chhetri2025application} showed that BERT outperformed CNN, LSTM, and HAN architectures in predicting the consequences of cyber attacks from the descriptions of CWE, but performance degraded under sparse and imbalanced conditions. VulnBERTa \cite{turtiainen2024vulnberta}, a hierarchical RoBERTa-based classifier trained on reports from the National Vulnerability Database (NVD), achieved strong accuracy on frequent CWE classes but significant degradation on rare labels. Contreras et al. \cite{contreras2023multiclass} evaluated BI-LSTM and BiGRU models on the Software Assurance Reference Dataset (SARD)  and the NVD with strong overall results but struggled to distinguish semantically overlapping CWE classes, highlighting the persistent barrier of long-tail label distributions.


\subsection{Semantic and Hierarchical Classification}
Previous work \cite{kota2024semantic} applied BERT-based cross-encoders with binary chaining to the CWE View-1003 hierarchy, improving accuracy on high-cardinality label sets but remaining limited in tail classes. V2W-BERT \cite{das2021v2w} demonstrated that label embeddings and textual CWE definitions improve representations for single-label CWE assignment. Classical ML approaches including Random Forests (RF) \cite{breiman2001random} and Support Vector Machines (SVM) \cite{hearst1998support} perform well on frequent weaknesses but consistently degrade on rare ones, motivating approaches that explicitly model label structure.


\subsection{Class Imbalance in Datasets}
SMOTE \cite{chawla2002smote} and ADASYN \cite{he2008adasyn} have proven effective for classifiers based on classical ML and CNN, and recent work has incorporated oversampling directly into deep learning training loops \cite{kishanthan2025deep}. However, the behavior of SMOTE and ADASYN on hierarchically structured CWE text embeddings, where parent-child constraints impose structural dependencies that linear interpolation cannot respect, remains largely unexplored, a gap this paper directly addresses. 

\section{Technical Background}
\label{sec:technical}

\subsection{Machine Learning Models}
\subsubsection{Random Forest}(RF)\cite{breiman2001random} constructs an ensemble of decision trees on bootstrapped subsets of training data, aggregating predictions to reduce variance and overfitting. Each tree recursively selects splits that maximize impurity reduction, performing well on structured feature spaces, but struggling on high-dimensional sparse text representations. Here, the best split is chosen as the feature and threshold that leads to the largest decrease in impurity between the parent and its children. 

\subsubsection{Support Vector Machine}(SVM) \cite{hearst1998support} learns a  maximum-margin hyperplane that separates classes in a high-dimensional feature space. A linear kernel is well suited for sparse text representations, making SVM a strong baseline for short-text classification tasks such as CWE descriptions. 

\subsection{Deep Learning Models}
\subsubsection{Convolutional Neural Networks}(CNN) \cite{yamashita2018convolutional} applies 1-D convolutional filters over token embeddings to capture local n-gram patterns, followed by max-pooling to retain the most salient activations. While effective for short range dependencies, CNNs are sensitive to perturbations in embedding space introduced by synthetic oversampling.

\subsubsection{Bidirectional Gated Recurrent Unit}(BiGRU)\cite{chung2014empirical} processes sequences in both forward and backward directions, concatenating hidden states at each position to capture full contextual dependencies. The gated mechanism controls information flow across time steps, making BiGRU effective for modeling order-sensitive vulnerability descriptions. 

\subsection{Transformers Based Models}
\subsubsection{Bidirectional Encoder Representations from Transformers}(BERT) \cite{devlin2019bert} uses stacked encoder layers with multi-head self-attention, pretrained via masked language modeling. Robustly Optimized Bidirectional Encoder Representations from Transformers (RoBERTa) \cite{liu2019roberta} improves on BERT by removing next sentence prediction, adopting dynamic masking, and training on substantially larger corpora, yielding stronger contextual representations across NLP benchmarks. SecureBERT \cite{aghaei2022securebert} extends RoBERTa through continuous pretraining on a large cybersecurity corpus and serves as the encoder backbone for our proposed model.
 
\section{Methodology }
\label{sec:methodology}

\subsection{Oversampling}
\begin{figure}
    \centering
    \includegraphics[width=\linewidth, height=4cm]{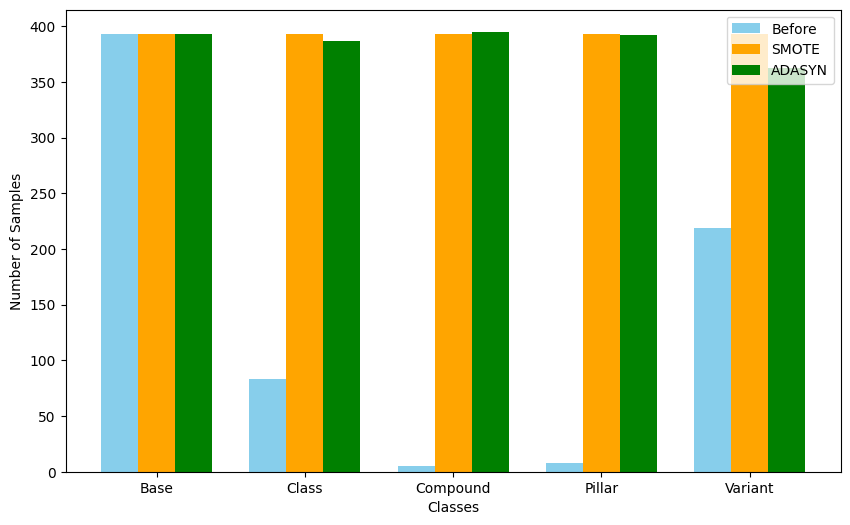}
    \caption{Class distribution before and after SMOTE and 
    ADASYN oversampling.}
    \label{fig:oversampling}
\end{figure}

The dataset exhibited severe class imbalance across five classes, i.e., \texttt{Base} (393), \texttt{Variant} (219), \texttt{Class} (83), \texttt{Compound} (8), and \texttt{Pillar} (5). Oversampling was applied exclusively to the training set to prevent data leakage. SMOTE equalized all classes to 393 samples using $k$-neighbors$=2$, while ADASYN produced near parity counts (\texttt{Compound} 395, \texttt{Base} 393, \texttt{Pillar} 392, \texttt{Class} 387, \texttt{Variant} 362) by concentrating synthesis on harder boundary regions. Figure \ref{fig:oversampling} illustrates the resulting  class distributions before and after resampling.

\subsection{Model Architectures}
\begin{figure*}
    \centering
    \includegraphics[width=\linewidth, height=4cm]{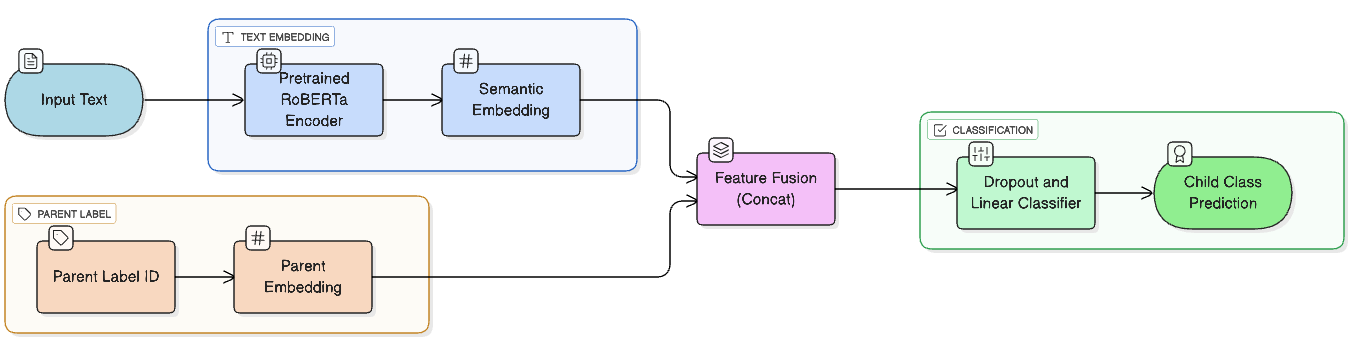}
    \caption{Overview of the Hierarchy-Aware RoBERTa framework.}
    \label{fig:hb_framewrok}
\end{figure*}

For classical baselines, RF was set with 100 decision trees ($n$-estimators $= 100$) and a fixed random seed (random-state $= 42$). The SVM classifier was configured to use a linear kernel, which is appropriate in high-dimensional, sparse feature representations.

For deep learning, CNN used 128 convolutional filters of kernel sizes [4,6,6] with \texttt{ReLU} activation and adaptive max pooling, while BiGRU used a hidden dimension of 128 with stacked recurrent layers to capture long-range contextual dependencies. Both models mapped tokens to 
256-dimensional embeddings trained end-to-end, fusing a 16-dimensional categorical status embedding via a 128-dimensional fusion layer (\texttt{ReLU}, dropout 0.2). BERT-base (uncased) used the \texttt{[CLS]} pooled output fused with a 16-dimensional status embedding, followed by 
a 128-dimensional linear layer (\texttt{ReLU}, dropout 0.2). All models were trained with a batch size of 8 for 15 epochs.

\subsection{Hierarchy-Aware RoBERTa}

We propose a Hierarchy-Aware RoBERTa architecture (Algorithm \ref{alg:hierarchy_aware_bert}) that explicitly incorporates label hierarchy information into Transformer-based text classification. 
A pretrained SecureBERT encoder\cite{aghaei2022securebert} produces a 768-dimensional 
\texttt{[CLS]} representation from the input text. To inject hierarchical knowledge, we introduce a learnable parent-category embedding that captures coarse-grained structural relationships among labels. Both parent categories are projected to a 24-dimensional dense embedding, trained together with the rest of the network.

The BERT pooled representation is then fused with the parent embedding, resulting in a unified feature representation. This represents both semantic content and hierarchical context. This fused representation is regularized using dropout with a rate of 0.3 before being passed to a linear classification head that outputs logits over the target classes.

\begin{algorithm}[t]

    \caption{Hierarchy-Aware RoBERTa for Text Classification}
    \label{alg:hierarchy_aware_bert}
    \begin{algorithmic}[1]
    \footnotesize
    \Require Input tokens $X = \{x_1, \dots, x_L\}$,
             attention mask $M$,
             parent category IDs $P$,
             pretrained RoBERTa encoder $\mathcal{B}$,
             parent embedding matrix $\mathbf{E}_p$
    \Ensure Class logits $\hat{y}$
    
    \State Encode input sequence using RoBERTa:
    \State \hspace{0.5cm} $\mathbf{H} \leftarrow \mathcal{B}(X, M)$
    \State Extract pooled representation:
    \State \hspace{0.5cm} $\mathbf{h}_{cls} \leftarrow \mathbf{H}_{[\text{CLS}]}$
    \State Obtain parent embedding:
    \State \hspace{0.5cm} $\mathbf{h}_p \leftarrow \mathbf{E}_p(P)$
    \State Fuse semantic and hierarchical representations:
    \State \hspace{0.5cm} $\mathbf{h}_f \leftarrow [\mathbf{h}_{cls}, \mathbf{h}_p]$
    \State Apply dropout:
    \State \hspace{0.5cm} $\tilde{\mathbf{h}}_f \leftarrow \text{Dropout}(\mathbf{h}_f)$
    \State Compute class logits:
    \State \hspace{0.5cm} $\hat{y} \leftarrow \mathbf{W}\tilde{\mathbf{h}}_f + \mathbf{b}$
    \State Optionally, obtain class probabilities and loss (training):
    \State \hspace{0.5cm} $\mathbf{p} \leftarrow \text{softmax}(\hat{y})$
    \State \hspace{0.5cm} $\mathcal{L} \leftarrow \text{CrossEntropy}(\mathbf{p}, y)$
    \State \Return $\hat{y}$
    \end{algorithmic}
\end{algorithm}

Figure \ref{fig:hb_framewrok} illustrates the proposed Hierarchy-Aware RoBERTa architecture, which combines semantic representations from a pretrained RoBERTa encoder \cite{aghaei2022securebert} with learned parent-label embeddings. The textual embedding and hierarchical context are fused via concatenation and passed through a dropout-regularized linear classifier to produce fine-grained child class predictions. This design enables hierarchy-informed classification by jointly leveraging contextual language understanding and structured label information. To ensure consistency across all models, deep learning models (CNN, BiGRU, BERT), and our proposed Hierarchy-Aware RoBERTa were all trained using the same batch size of 8 and 15 epochs.

\section{Experimental Procedure}
\label{sec:experiment}

\subsection{Dataset}
\begin{figure}[!t]
    \centering
    \includegraphics[width=\linewidth, height=4cm]{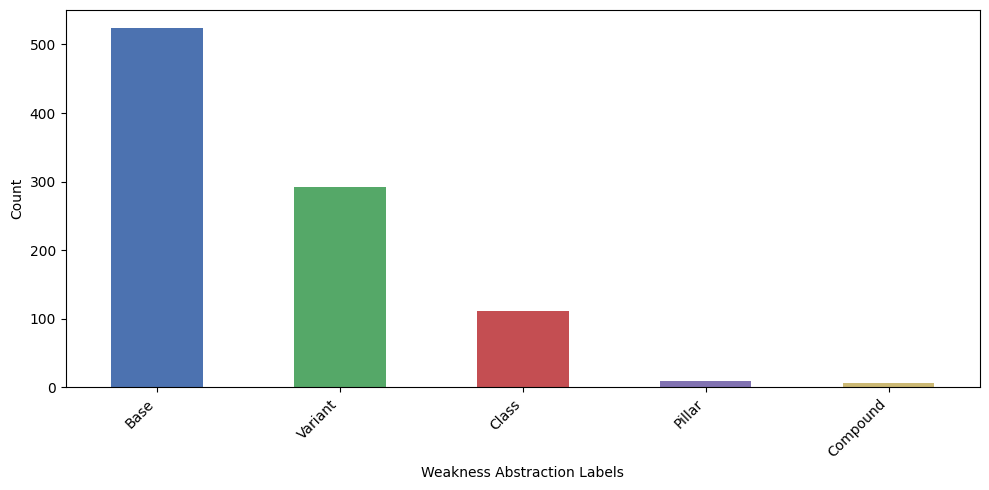}
    \caption{Frequency of  Weakness Abstraction Labels.}
    \label{fig:dataset}
\end{figure}

The dataset used in this study was sourced from an enhanced version of the MITRE Common Weakness Enumeration (CWE) dataset. The CWE repository is continuously updated and maintained to incorporate newly identified software vulnerabilities and attack patterns. The dataset\footnote{https://cwe.mitre.org/data/slices/1000.html}  contains \texttt{944} data points with  \texttt{23} columns. For the experiments, we selected three columns (i.e., \texttt{CWE-ID, Name, Weakness Abstraction, Description}). The \texttt{Weakness Abstraction} has five unique values as presented in Figure \ref{fig:dataset}. The dataset includes five classes in ``$Weakness Abstraction$'' column, Base sample of 524, Variant sample of 292, Class sample of 111, Pillar sample of 10, and a Compound sample of 7. 

\subsection{Data Processing}

\begin{figure}[htb]
    \centering
    \includegraphics[width=\linewidth, height=3cm]{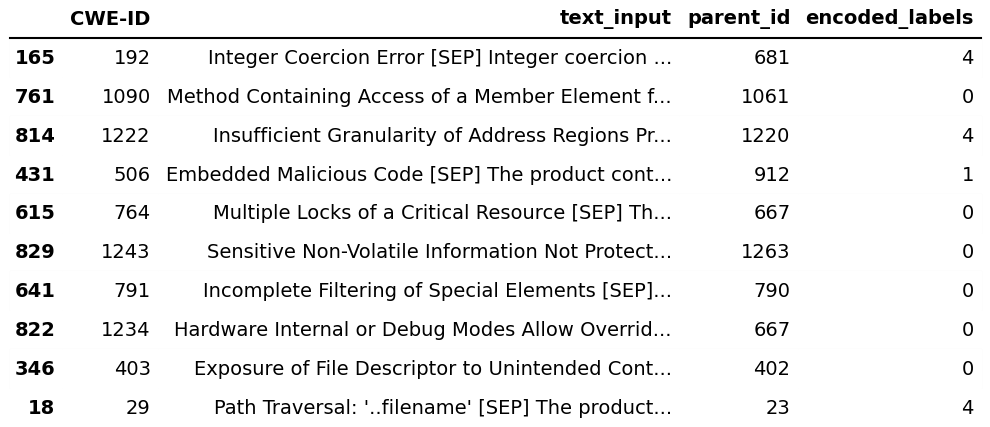}
    \caption{A sample of CWE dataset after data preprocessing}
    \label{fig:processed_data}
\end{figure}

In this experiment, all five abstraction classes were retained for classification. The Name and Description fields were concatenated using a \texttt{[SEP]} token after removing noise patterns and trimming whitespace. Figure~\ref{fig:processed_data} illustrates a sample of the processed dataset. Labels from \texttt{Weakness Abstraction} were encoded as integers (0--4) and tokenized using BertTokenizer. Data were split 75:25 for training and testing, with all inputs padded or truncated to a maximum sequence length of 128 tokens.

\begin{figure}[!t]
    \centering
    \includegraphics[width=\linewidth, height=5cm]{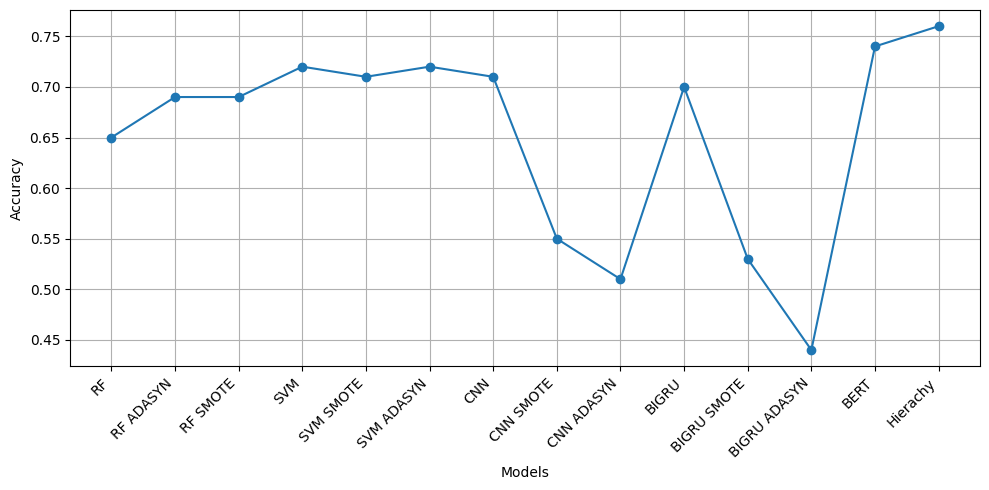}
    \caption{Model Accuracy Comparison Across Models}
    \label{fig:accuracy}
\end{figure}

\section{Results}
\label{sec:results}

\subsection{Model Accuracy}

Figure \ref{fig:accuracy} compares the performance of various  machine learning, deep learning, and transformer-based models in terms of their accuracy as the main metric. 
RF achieved a baseline accuracy of 0.65, improving its performance to 0.69 with both SMOTE and ADASYN. SVM performed better than RF with an accuracy of 0.72, while SMOTE and ADASYN demonstrated the same performance, i.e., 0.71 and 0.72, respectively. Such findings suggest that oversampling is weakly helpful with classical models and helps address class imbalance.
Similarly, the vanilla CNN dropped from 0.71 to 0.55 with SMOTE and 0.51 with ADASYN, 
while BiGRU fell from 0.70 to 0.53 and 0.44, respectively, confirming that oversampling can introduce noise that adversely impacts sequence-based and convolutional frameworks. BERT achieved an accuracy of 0.74, while the proposed Hierarchy-Aware RoBERTa achieved the highest accuracy of \textbf{0.76}, demonstrating the advantage of structural priors over synthetic data generation.

\subsection{Classification Performance }
\vspace{0.15cm}

\begin{table*}[ht]

    \centering
    \caption{Per-Class Performance of Models Across Data Augmentation Methods}
    \label{tab:class-metrics}
\resizebox{\linewidth}{!}{
    \small
    \begin{tabular}{l|l|lccccccccc}
    \toprule
    \multirow{2}{*}{\textbf{Category}} & 
    \multirow{2}{*}{\textbf{Model}} &
    \multirow{2}{*}{\textbf{Class}} &
    \multicolumn{3}{c}{\textbf{BASELINE}} & 
    \multicolumn{3}{c}{\textbf{SMOTE}} &
    \multicolumn{3}{c}{\textbf{ADASYN}}  \\
    \cmidrule(lr){4-6}
    \cmidrule(lr){7-9}
    \cmidrule(lr){10-12}
    
    & & & Precision & Recall & F1 & Precision & Recall & F1 & Precision & Recall & F1 \\
    \midrule
    
    \multirow{10}{*}{ML}  
    & \multirow{5}{*}{RF} 
        & Base     & 0.63 & 0.95 & 0.76     & 0.68 & 0.89 & 0.77    & 0.70 & 0.86 & 0.77   \\
    &   & Class    & 0.50 & 0.04 & 0.07     & 0.60 & 0.32 & 0.42    & 0.67 & 0.43 & 0.52     \\
    &   & Compound & 0.00 & 0.00 & 0.00     & 0.00 & 0.00 & 0.00    & 0.00 & 0.00 & 0.00   \\
    &   & Pillar   & 0.00 & 0.00 & 0.00     & 0.00 & 0.00 & 0.00    & 0.00 & 0.00 & 0.00    \\
    &   & Variant  & 0.77 & 0.37 & 0.50     & 0.76 & 0.51 & 0.61    & 0.70 & 0.53 & 0.60   \\
    & &\parbox[c]{1.5cm}{\centering \bf{Weighted}\\\bf{Average}} 
    & \bf{0.65} & \bf{0.65} & \bf{0.58} 
    & \bf{0.68} & \bf{0.69} & \bf{0.67} 
    & \bf{0.68} & \bf{0.69} & \bf{0.68} 
     \\
   
    \cmidrule(lr){2-12}
    
    & \multirow{5}{*}{SVM} 
        & Base     & 0.74 & 0.85 & 0.79 & 0.73 & 0.84 & 0.78 & 0.74 & 0.85 & 0.79  \\
    &   & Class    & 0.50 & 0.46 & 0.48 & 0.50 & 0.46 & 0.48 & 0.50 & 0.46 & 0.48  \\
    &   & Compound & 0.00 & 0.00 & 0.00 & 0.00 & 0.00 & 0.00 & 0.00 & 0.00 & 0.00    \\
    &   & Pillar   & 0.00 & 0.00 & 0.00 & 0.00 & 0.00 & 0.00 & 0.00 & 0.00 & 0.00    \\
    &   & Variant  & 0.78 & 0.63 & 0.70 & 0.76 & 0.62 & 0.68 & 0.78 & 0.63 & 0.70   \\
     & &\parbox[c]{1.5cm}{\centering \bf{Weighted}\\\bf{Average}} 
    & \bf{0.71} & \bf{0.72} & \bf{0.71} & \bf{0.70} & \bf{0.71} & \bf{0.70} 
    & \bf{0.71} & \bf{0.72} & \bf{0.71} 
    \\
    \midrule
    
    \multirow{15}{*}{DL}  
    & \multirow{5}{*}{CNN}
        & Base & 0.70 & 0.92 & 0.79 & 0.70 & 0.70 & 0.70 & 0.70 & 0.66 & 0.68    \\
    &   & Class & 0.56 & 0.18 & 0.27 & 0.14 & 0.11 & 0.12 & 0.09 & 0.07 & 0.08  \\
    &   & Compound & 0.00 & 0.00 & 0.00 & 0.02 & 0.50 & 0.04 &0.02 & 0.50 & 0.04     \\
    &   & Pillar & 0.00 & 0.00 & 0.00 &0.00 & 0.00 & 0.00 & 0.00 & 0.00 & 0.00   \\
    &   & Variant & 0.79 & 0.58 & 0.67 &0.90 & 0.48 & 0.62 & 0.79 & 0.45 & 0.57     \\
     & &\parbox[c]{1.5cm}{\centering \bf{Weighted}\\\bf{Average}} 
    & \bf{0.70} & \bf{0.71} & \bf{0.68} & \bf{0.68} & \bf{0.56} & \bf{0.60} 
    & \bf{0.64} & \bf{0.52} & \bf{0.56}     \\
    \cmidrule(lr){2-12}
    
    & \multirow{5}{*}{BI-GRU}
        & Base & 0.74 & 0.80 & 0.77 & 0.81 & 0.56 & 0.67 & 0.77 & 0.41 & 0.54   \\
    &   & Class & 0.50 & 0.32 & 0.39 & 0.19 & 0.14 & 0.16 & 0.22 & 0.25 & 0.23   \\
    &   & Compound & 0.00 & 0.00 & 0.00 & 0.00 & 0.00 & 0.00 & 0.02 & 0.50 & 0.04    \\
    &   & Pillar & 0.00 & 0.00 & 0.00 & 0.11 & 1.00 & 0.20 & 0.06 & 0.50 & 0.11   \\
    &   & Variant & 0.70 & 0.73 & 0.71 & 0.61 & 0.63 & 0.62 & 0.66 & 0.58 & 0.61  \\
     & &\parbox[c]{1.5cm}{\centering \bf{Weighted}\\\bf{Average}} 
    & \bf{0.68} & \bf{0.71} & \bf{0.69} & \bf{0.66} & \bf{0.53} & \bf{0.58} 
    & \bf{0.66} & \bf{0.44} & \bf{0.52} 
     \\
    \cmidrule(lr){2-12}
    
    & \multirow{5}{*}{BERT}
        & Base & 0.82 & 0.75 & 0.78     & - & - & -    & - & - & -  \\
    &   & Class & 0.57 & 0.43 & 0.49    & - & - & -    & - & - & -   \\
    &   & Compound & 0.00 & 0.00 & 0.00     & - & - & -    & - & - & -    \\
    &   & Pillar & 0.00 & 0.00 & 0.00   & - & - & -    & - & - & -    \\
    &   & Variant & 0.69 & 0.89 & 0.78  & - & - & -    & - & - & -   \\
     & &\parbox[c]{1.5cm}{\centering \bf{Weighted}\\\bf{Average}} 
    & \bf{0.74} & \bf{0.74} & \bf{0.73} & 
    \bf{-} & \bf{-} & \bf{-} 
    & \bf{-} & \bf{-} & \bf{-} 
    \\
     \midrule
    
    \multirow{10}{*}{Ours}  
    & \multirow{5}{*}{Hierarchy-Aware RoBERTa}
        & Base & 0.78 & 0.85 & 0.81     & - & - & -    & - & - & -  \\
    &   & Class & 0.56 & 0.64 & 0.60    & - & - & -    & - & - & -  \\
    &   & Compound & 0.00 & 0.00 & 0.00     & - & - & -    & - & - & - \\
    &   & Pillar & 0.00 & 0.00 & 0.00   & - & - & -    & - & - & -  \\
    &   & Variant & 0.84 & 0.76 & 0.76  & - & - & -    & - & - & -  \\
     & &\parbox[c]{1.5cm}{\centering \bf{Weighted}\\\bf{Average}} 
    & \bf{0.76} & \bf{0.76} & \bf{0.76 } 
    &\bf{-} & \bf{-} & \bf{-} 
    & \bf{-} & \bf{-} & \bf{-} \\

    \bottomrule
    \end{tabular}
    }
\end{table*}

Table~\ref{tab:class-metrics} reports the precision, recall, and F1-score per-class across all models and augmentation conditions.
The improvement was primarily observed in minority classes. For example, the \texttt{Class} F1-score 
increased from 0.07 to 0.42 with SMOTE and 0.52 with ADASYN for RF. However, \texttt{Compound} and \texttt{Pillar} remained at 0.00 F1 across all settings. SVM showed stronger overall performance with a weighted F1 of 0.71, maintaining stable scores of 0.70-0.71 under both oversampling methods, 
indicating robustness on well-represented classes.

For deep learning models, CNN achieved a baseline weighted F1 of 0.68, with strong performance on \texttt{Base} (F1$=0.79$) and \texttt{Variant} (F1$=0.67$). However, applying SMOTE and ADASYN reduced the performance to 0.60 and 0.56, respectively, with \texttt{Class} F1 dropping to as 
low as 0.08 under ADASYN. BiGRU achieved a baseline weighted F1 of 0.69, dropping to 0.58 with SMOTE and 0.52 with ADASYN, with substantial degradation in minority classes, highlighting the sensitivity of sequence-based models to oversampling-induced noise.

Transformer-based BERT achieved a strong baseline weighted F1 of 0.73, driven by high performance on \texttt{Base} (F1$=0.78$) and \texttt{Variant} (F1$=0.78$). No augmentation was applied for BERT, as its contextual embeddings were sufficient without synthetic oversampling. Our Hierarchy-Aware RoBERTa further improved performance, achieving the highest weighted F1 of \textbf{0.76}, 
consistently outperforming all models across \texttt{Base} (F1$=0.81$), \texttt{Class} (F1$=0.60$), and \texttt{Variant} (F1$=0.76$). In particular, \texttt{Compound} and \texttt{Pillar} remain at 0.00 F1 in all models, reflecting the persistent challenge of supervised learning under extreme 
data sparsity.

\section{Discussion}
\label{sec:discussion}

Although data augmentation techniques such as SMOTE and ADASYN are widely used to address class imbalance in classical machine learning, they are not well suited for transformer-based models. BERT does not use fixed, low-dimensional feature vectors, but learns contextualized, high-dimensional semantic representations directly from raw text, meaning that synthetic interpolation produces samples that lack linguistic coherence and add noise rather than signal.  
For CNN and BiGRU models, small perturbations in embedding space correspond to large taxonomic shifts, causing models to overfit to synthetic artifacts and degrading generalization 
performance on minority and hierarchical classes.

Furthermore, SMOTE and ADASYN do not conserve the hierarchical relationship that exists within the data. Classes such as \texttt{Variant} are semantically dependent on their parent classes (e.g., \texttt{Base}). Synthetic oversampling does not respect such constraints and can produce samples that are inconsistent with their underlying class structure and confuse the model during training. 
In the CWE taxonomy, each entry links to a \texttt{Related Weakness} via an explicit parent-child chain, meaning errors at higher abstraction levels cascade downward to finer-grained classes, a structural dependency that oversampling methods ignore entirely by treating each sample 
as an independent entity in a flat feature space. CWEs are not isolated texts but are organized as a tree structured graph where more abstract weaknesses act as semantic parents to narrower ones, and interpolating between embeddings from different parent nodes produces representations of 
non-existent or indistinct weaknesses, giving statistically balanced but structurally invalid embeddings. 

Our Hierarchy-Aware RoBERTa model explicitly incorporates structural knowledge by introducing a learnable parent embedding. In the CWE taxonomy, weakness abstractions are strictly hierarchical.  For instance, a \texttt{"Variant"}  class is semantically and structurally dependent on its corresponding  \texttt{"Base"} parent. To model this dependency, we formulate the classification task not as an independent prediction $P(Y_{child}|X_{text})$, but as a conditional probability problem $P(Y_{child}|X_{text}, Y_{parent})$. We utilize \texttt{"Parent ID"} as a structural prior to guide the model to learn that if the parent is a \texttt{Base} type, the child is likely a \texttt{Variant}. We define two input vectors: Text Vector ($V_{text}$): The output from BERT (768 dimensions). Parent Vector ($V_{graph}$): A learnable embedding for the parent ID (24 dimensions). The model fuses them via concatenation:$V_{final} = \text{Concat}(V_{text}, V_{graph})$, $V_{final} \in \mathbb{R}^{768 + 24}$. The classifier then predicts based on this combined knowledge: $y_{pred} = \text{Softmax}(W \cdot V_{final} + b)$. 



\subsection{Limitations of the Hierarchy-Aware RoBERTa Model}
Despite its improved performance, the proposed Hierarchy-Aware RoBERTa model has several limitations. 
First, the architecture assumes parent-level metadata availability during inference, though this is consistent with coarse-to-fine vulnerability pipelines where high-level categories are readily identifiable via static analysis tools (SAST) or manual triage. Second, the hierarchy is treated as fixed, and the model may not generalize to new or rearranged nodes without retraining, as the CWE taxonomy is subject to continual updates. Third, only direct parent information is captured, 
disregarding multi-hop ancestors, sibling relationships, and tree depth, which may limit discrimination among deeply nested classes such as \texttt{Compound} and \texttt{Pillar}. When training data is virtually absent, hierarchy awareness alone is insufficient, suggesting that weak supervision or external knowledge may be necessary. Finally, the learnable parent embeddings add 
modest parameter overhead that could hinder scaling to large taxonomies, and the approach has been evaluated only within the CWE domain, leaving generalizability to other hierarchical tasks empirically unvalidated.

\section{Conclusion}
\label{sec:conclusion}

This study examined the effectiveness of oversampling techniques and hierarchy-aware modeling to address class imbalance in CWE vulnerability classification. Through extensive experimentation, we demonstrated that commonly used oversampling methods, such as SMOTE and ADASYN, provide limited benefits for machine learning models (i.e., RF and SVM) and consistently degrade performance for deep learning models (i.e., CNN and BiGRU). These methods generate synthetic samples through linear interpolation in embedding space, an assumption that does not hold for contextualized text representations and often disrupts the semantic and hierarchical integrity of CWE labels.

To overcome these limitations, we proposed a Hierarchy-Aware RoBERTa model that explicitly integrates parent–child relationships from the CWE taxonomy into the classification process. By incorporating learnable parent embeddings alongside contextual language representations, the model enforces structural consistency and improves discrimination among minority and semantically dependent classes. The proposed model achieves a weighted F1-score of $0.76$ without data augmentation, outperforming all baselines, including BERT, which achieved $0.74$, demonstrating that structural priors are more effective than synthetic data generation in hierarchical imbalanced cybersecurity datasets.

Despite these gains, extremely sparse categories such as \texttt{Compound} and \texttt{Pillar} remain difficult to predict, highlighting the inherent limitations of supervised learning when training data is severely limited. Future research may extend the hierarchy-aware framework to incorporate multi-hop ancestry and graph neural networks over the full CWE hierarchy, explore weak supervision and 
label propagation for extremely sparse classes, and evaluate the proposed approach across other hierarchical cybersecurity datasets to establish its generalizability beyond the CWE domain.

\section*{Acknowledgment}
This research is partially supported by the U.S. National Science Foundation (Award \#: 2319802).

\bibliography{refs}{}
\bibliographystyle{plain}

\end{document}